\def\year{2021}\relax
\documentclass[letterpaper]{article} % DO NOT CHANGE THIS
\usepackage{aaai21}  % DO NOT CHANGE THIS
\usepackage{times}  % DO NOT CHANGE THIS
\usepackage{helvet} % DO NOT CHANGE THIS
\usepackage{courier}  % DO NOT CHANGE THIS
\usepackage[hyphens]{url}  % DO NOT CHANGE THIS
\usepackage{graphicx} % DO NOT CHANGE THIS
\usepackage{natbib}  % DO NOT CHANGE THIS AND DO NOT ADD ANY OPTIONS TO IT
\usepackage{caption} % DO NOT CHANGE THIS AND DO NOT ADD ANY OPTIONS TO IT
\usepackage{balance} 
\usepackage{amsmath}
\usepackage{mathtools}
\usepackage{amsfonts}
\usepackage{amsthm}

\usepackage{times}
\usepackage{soul}
\usepackage{url}

\usepackage{graphicx}
\usepackage[inline]{enumitem}
\newcommand{\parg}[1]{\vspace{1mm}\noindent\textbf{#1.}\hspace{2mm}}

\title{Business Entity Matching with Siamese Graph Convolutional Networks}

\author{
Evgeny Krivosheev$^{1,2}$,
Mattia Atzeni$^{1,3}$,
Katsiaryna Mirylenka$^{1}$\\
Paolo Scotton$^1$,
Christoph Miksovic$^1$ and
Anton Zorin$^1$
    \\
}
\affiliations{
    $^1$ IBM Research - Zurich, Switzerland,\\
    $^2$ University of Trento, Italy\\
    $^3$ EPFL, Switzerland\\
    evgeny.krivosheev@unitn.it,
    \{atz, kmi, psc, cmi, zor\}@zurich.ibm.com
}

\begin{document}

\maketitle

\begin{abstract}
Data integration has been studied extensively for decades and approached from different angles. However, this domain still remains largely rule-driven and lacks universal automation.
Recent developments in machine learning and in particular deep learning have opened the way to more general and efficient solutions to data-integration tasks.
In this paper, we demonstrate an approach that allows modeling and integrating entities by leveraging their relations and  contextual information.
This is achieved by combining siamese and graph neural networks to effectively propagate information between connected entities and support high scalability.
We evaluated our approach on the task of integrating data about business entities, demonstrating that it outperforms both traditional rule-based systems and other deep learning approaches. 
\end{abstract}

\section{Introduction}
\label{sec:introduction}

Although knowledge graphs (KGs) and ontologies have been exploited successfully for data integration~\cite{Trivedi2018,Azmy2019}, entity matching involving structured and unstructured sources has usually been performed by treating records without explicitly taking into account the natural graph representation of structured sources and the potential graph representation of unstructured data~\cite{Mudgal2018,gschwind2019}. 
%Implicit graph representations of database records have only been exploited for record-linkage tasks by considering the similarity of the attributes between a query and a candidate record.
% In this way, implicit connections between attributes are taken into account.
%Similarities between records are usually computed by means of multi-criteria scoring without considering the direct attribute for information flow~\cite{gschwind2019}. 
%\begin{figure}[tb]
%	\begin{center}
%		\includegraphics[width=0.4\textwidth]{img/entity_matching_example.pdf}  
%		\caption{Example of entity matching and graph modeling}
%		\label{fig:linkage-example}
%	\end{center}
%	\vspace*{-20pt}	
%\end{figure}
To address this limitation, we propose a methodology for leveraging graph-structured information in entity matching.
%The main idea consists of exploiting KGs to create distributed representations of nodes, so that entities that are linked in the KG are also close in the embedding space.
Recently, graph neural networks (GNNs) yielded promising results for building and processing distributed representations of nodes in a graph. This kind of approach has already achieved state-of-the-art performance on several tasks, including link prediction, node classification, and graph generation~\cite{Zhang2018,Ying2018,Xu2019,Chami2019}. 
The key idea of GNNs is to propagate useful information from a node to its neighbors, allowing us to build discriminative node embeddings from the \textit{data} and \textit{their structure}.  This property makes GNNs extremely suited to relational databases as well.
In particular, databases that store information about business entities explicitly or implicitly incorporate relations between companies and their attributes (e.g., in terms of ownership, branches, and subsidiaries)~\cite{mirylenka2019}. 
%This means that companies and their connections can be naturally represented as a KG.
%We built KGs from the database at hand about companies and their relations (e.g., in termsof ownership and subsidiaries).
%This motivated us to extract the KG from our database of companies and their relations (e.g., in terms of ownership and subsidiaries).
We therefore envision that modeling hidden representations of business entities with GNNs will improve performance of entity matching in this use case and, more generally, in the field of data integration.

%\todo[inline]{almost always a db can be modeled as a graph unless it's completely denormalized, but even in that case it can be. should we skip this paragraph?}
%As a simple example,  Figure ~\ref{fig:linkage-example} depicts graph modeling from a reference database, and shows how company entities, extracted from news articles, are linked to corresponding records.

To investigate this hypothesis, this paper presents a graph-based entity-matching system that is capable of learning the distance function between business entities using a model based on Siamese Graph Convolutional Networks (S-GCN)~\cite{krivosheev2020}.
The neural network is optimized to produce small distances for nodes belonging to the same business entity and large distances for unrelated nodes.
We show the power and flexibility of our S-GCN model in recognizing company names with typos, automatically extracting unseen abbreviations, and performing a semantic search on business entities in a relational database.

%Therefore, the trained network can then be used to assess the similarity between a query and any reference node. S-GCNs are especially useful when the task involves a large number of classes, but only a few examples for each label are available.

\section{Siamese GCN for Entity Matching}
\label{sec:siamese_GCN_for_record_linkage}
\begin{figure}[t]
	%\vspace{-7pt}
	\begin{center}
		\includegraphics[width=0.87\linewidth]{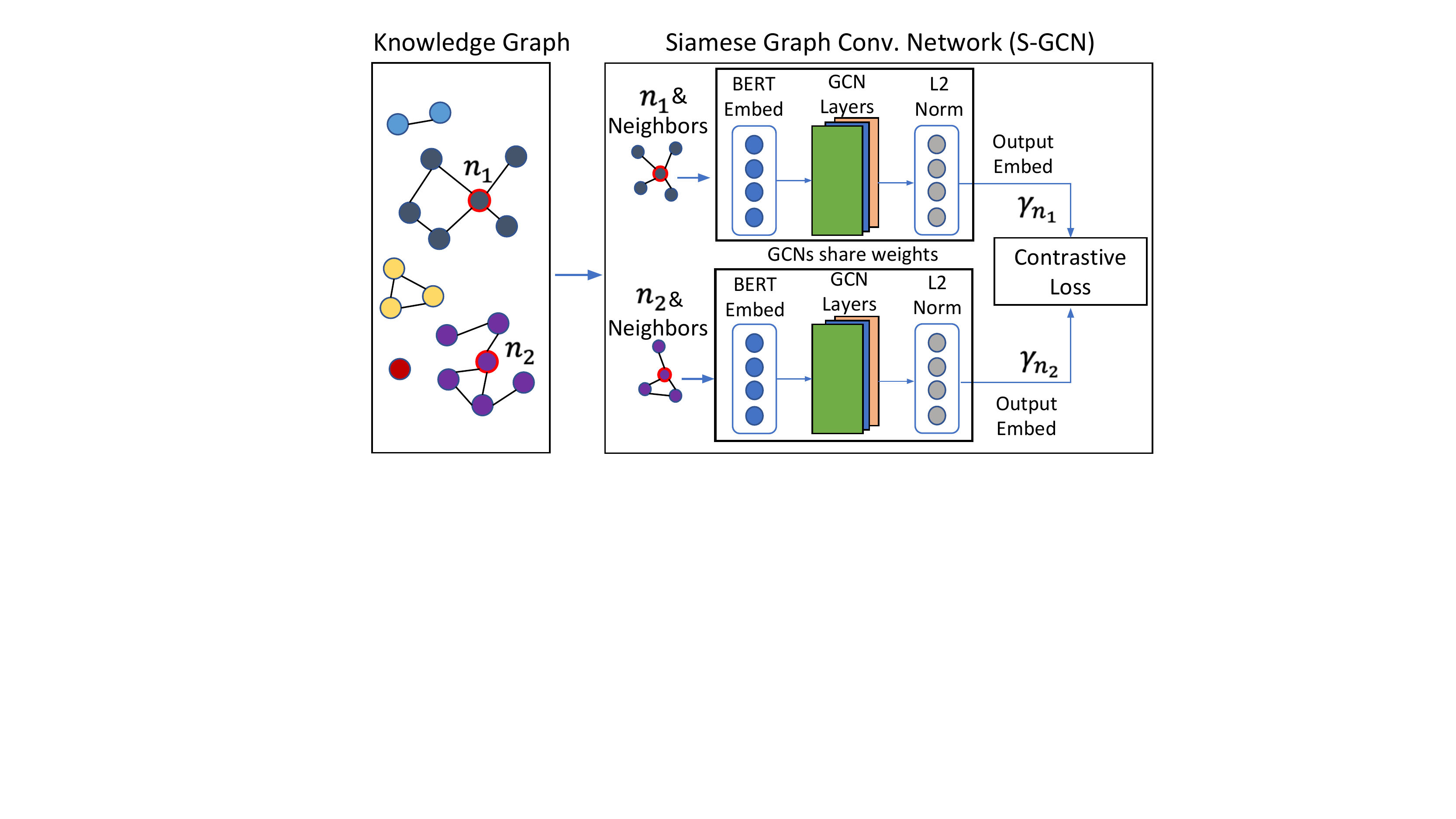}  
		\caption{Architecture of the Siamese Graph Conv Network}
		\label{fig:sgcn}
	\end{center}
	\vspace{-17pt}
\end{figure}

\vspace{-3pt}
\parg{Network architecture} 
We propose a model architecture that combines the advantages of graph convolutional networks (GCNs)~\cite{Kipf2017} and siamese networks~\cite{Bromley1993} to address the entity-matching task. GCNs are a type of graph neural network that shares filter parameters among all the nodes, regardless of their location in the graph.
Our Siamese Graph Convolutional Network (S-GCN) incorporates two identical GCNs, as shown in Figure \ref{fig:sgcn}.
The primary objective of our S-GCN architecture is to learn discriminative embeddings of nodes in a knowledge graph, in such a way that they can then be used for entity matching with previously unobserved data.
In our use case, the input layer of the network expects features extracted from textual attributes of the nodes using a pre-trained BERT model \cite{Devlin2019}. Next, some GCN layers followed by $L_2$ normalization are used to produce the output embedding for each node.
During the training process, the first GCN focuses on a given node $n_1$ and its local neighborhood in the KG, to produce an $M$-dimensional embedding $\gamma_{n_1} \in \mathbb{R}^M$ of node $n_1$. Similarly, the same procedure is repeated with the second GCN for a different node $n_2$, to create an embedding $\gamma_{n_2} \in \mathbb{R}^M$ for $n_2$. Our siamese network is optimized with a \textit{contrastive  loss}  function~\cite{Hadsell2006} between the outputs of the two GCNs, namely $\gamma_{n_1}$ and $\gamma_{n_2}$. Intuitively, this loss will be small for two nodes that are connected in the graph, whereas it will be larger for nodes that are not related to each other. We make use of \textit{Subspace Learning}~\cite{Wang2017} to carefully and efficiently select informative pair of nodes at training time. 

\parg{Entity matching with S-GCN} 
The network is trained on a reference KG (which comprises business entities in our use case) so that the structure of the graph guides the training process. After the S-GCN model has been trained, it projects the KG into an embedding space, where nodes that represent the same entity are mapped to similar vectors. At test time, the trained model can be used to project an unseen entity mention into the same embedding space. Then, $K$-NN search can applied to retrieve the closest business entity.

\section{System Evaluation}
\parg{Data} 
\label{sec:graph_extraction}
We set up a data-integration task on a proprietary relational database that contains fine-grained information about business entities (i.e. companies). Companies represented in the database may have different branches, subsidiaries and headquarters, which can be grouped together hierarchically into a tree that represents a single entity.
Therefore, we can model our reference database as a graph composed of several disjoin trees.
Each node in the graph has a textual attribute that provides the precise name of the branch associated to that node.
%Each company entry represented by  business location, industry, a textual name, and, optionally, up to three alternative names.
%We are interested in linking a given company name to a company in the reference database.
%To this end, we represent families in the database as a graph, where each node is associated to the name of a company.
%Relations between nodes are manually derived from the hierarchical representation of company families.
%A relation between two nodes exists if they are adjacent business entities in the   
%hierarchy defined by the database.
The resulting training graph contains approximately $40k$ nodes organized into $1.7k$ business entities.
As a data-augmentation step, we generate an additional canonical or normalized version of a company name and link it to the real name in the graph, using conditional random fields, as described in \cite{gschwind2019}. This step yields an enriched training graph with $70k$ nodes. 
We are interested in linking unseen company names to an entity in the reference database, and we evaluate our proposed approach using a test set with $1.8k$ company names.

\parg{Scalability}
 The linkage task involves searching for the nearest-neighbors of a vector in $\mathbb{R}^M$. This is a computationally difficult task that has resulted in the development of \textit{approximate nearest-neighbors} (ANN) algorithms. \cite{aumuller2017}~have evaluated a set of state-of-the-art ANN algorithms under different conditions on a variety of datasets. They show that it is possible to retrieve ten nearest-neighbors (Euclidean distance) in less than 100 $\mu$s in a 960-dimensional embedding space with recall of 1.00 on servers with Intel Xeon Platinum 8124M CPUs.

%To investigate the feasibility of the approach, we set up a data integration task on a proprietary relational database that contains fine-grained information about business entities. 
%Each record in the business entity 
% database at hand corresponds to a different company business location and is assigned a  unique local identifier
%that can be used to track  business entities precisely.
%Company entities represented in the database include different branches, subsidiaries and headquarters, each of which reports directly or indirectly to a given global identifier. 
%Hence, companies related to the same global identifier can be grouped together and represented hierarchically as a single family.
%In addition, for each company entry,  
%we have a textual name, and, optionally, up to three alternate names.
% 
%We are interested in linking a company name to a family in the  
% database. To this end, we represent families in the 
%  database as KG, where each node is a company name.
%Relations between nodes are manually derived from the hierarchical representation of 
%company families and from the aliases provided in the database.
%More precisely, a relation between two nodes exists if they are adjacent business entities in the   
%hierarchy defined by the database, or if they are aliases of each other.
%
%
%We generate an additional short or normalized versions of a company name and link it to the real name, using conditional random fields (CRFs) as described in \cite{gschwind2019}.

\parg{Experiments} 
We compared our S-GCN model against three baselines, namely \textit{(i)} a record-linkage system (RLS) designed for company entities~\cite{gschwind2019}, \textit{(ii)} a feed-forward neural network (NN), and \textit{(iii)} a model based on graph convolutional networks (GCN). Both the GCN and NN models use BERT features as input and a softmax output layer.
We evaluated these algorithms in terms of their accuracy on the test set and we report experimental results in Table \ref{tab:results}.
As we can see, our S-GCN model consistently outperforms the baselines both on the original KG and on the data augmented with artificially generated canonical names~\cite{gschwind2019}. Moreover, performing data augmentation on the KG greatly improves the performance of both models based on graph convolutions.
Adding such augmented names allows S-GCN to effectively propagate information from the most descriptive parts of the original names. 
We also performed some experiments using our S-GCN model with fine-tuned BERT embeddings. The fine-tuning allowed us to largely improve the performance of the final system, reaching an accuracy of 0.90 on the test set.

\begin{table}[t]
	%\vspace*{-8.5pt}	
	%\vspace*{-13pt}	
	\begin{center}
		\scalebox{0.85}
		{
			\begin{tabular}{|l|c|c|}
				\hline
				\textbf{Algorithm}& \textbf{KG w/o Augmentation} & \begin{tabular}[c]{@{}l@{}} \textbf{Augmented KG} \end{tabular} \\ \hline
				\textbf{RLS} & 0.71 & 0.78 \\ \hline
				\textbf{NN} & 0.71 & 0.75 \\ \hline
				\textbf{GCN} & 0.56 & 0.70 \\ \hline
				\textbf{S-GCN} & \textbf{0.75} & \textbf{0.85} \\ \hline  
			\end{tabular}
		}
	\end{center}
	\vspace*{-6pt}	
	\caption{Accuracy of entity matching on the test set}
	\label{tab:results}
	\vspace*{-15pt}	
\end{table}

%\begin{table}[tbh]
%	\vspace*{-8.5pt}	
%	\caption{Accuracy of entity matching on unseen data}
%	\vspace*{-13pt}	
%	\label{tab:res}
%	\begin{center}
%	\scalebox{0.85}
%	{
%		\begin{tabular}{|l|c|c|}
%			\hline
%			\textbf{Algorithm}& \textbf{Swiss dataset} & \begin{tabular}[c]{@{}l@{}} \textbf{Swiss dataset} \\  \textbf{+ short names} \end{tabular} \\ \hline
%			\textbf{RLS} & 0.71 & 0.78 \\ \hline
%			\textbf{Perceptron} & 0.71 & 0.75 \\ \hline
%			\textbf{S-GCN} & \textbf{0.86} & \textbf{0.90} \\ \hline  
%		\end{tabular}
%	}
%	\end{center}
%	\vspace*{-15pt}	
%\end{table}

%Analyzing the results on the Swiss dataset (Table~\ref{tab:res}), we can easily see that enriching the graphs with short names allows improving accuracy by a large margin. Most notably, for S-GCN, accuracy rises by 13\%. 
%Adding short names increases the variability of training data and allows GNNs to effectively propagate information from short versions to the original names. 
%Thus, the usage of the higher number of alias generating techniques might be beneficial.

\section{Demonstration}
\label{sec:demonstration}

%\begin{figure}[b]
%	\vspace{-10pt}
%	\centering
%	\includegraphics[width=.85\columnwidth]{img/ui.png}
%	\caption{Interface of the deployed entity-matching service}
%	\label{fig:ui}
%\end{figure}
We deployed the network as a service on Kubernetes and we built a user interface to easily search company names and retrieve the match provided by our model.
The demonstration will assess the robustness of the system and showcase some interesting properties learned implicitly by our S-GCN architecture. As a first example, the neural network learned to recognize some abbreviations of company names. Both the names \emph{``IBM''} and \emph{``International Business Machines''} are matched correctly to the same company, as shown in. Moreover, we applied different kinds of perturbations to company names (e.g., swapping, inserting, replacing, or removing characters) and we noticed that the system was still able to retrieve the correct entity. This shows that the model has implicitly learned an appropriate string similarity measure for our use case. As an example of such ability, we mention that some wrong input names like \emph{``Glecor''}, \emph{``lgencore''}, \emph{``AGlencore''}, \emph{``Glenocre''}, and \emph{``Glencorh''} were all matched correctly to the most similar company in the KG, namely \emph{``Glencore PLC''}.
In some cases, when the number of typos introduced in the input name was large enough to deviate the prediction of the system, the model still provided a company similar to the input text.
Moreover, we also observed that the system was able to perform a semantic search on the reference database. For instance, if provided with the input word \emph{``food''},  the model retrieves the company \emph{``Valentino Gastronomie AG''}, which operates in the food domain.
Similarly, the word \emph{``communications''} is matched to the company \emph{``Schweizerische Radio- und Fernsehgesellschaft''}.
This ability is probably given by high-level semantic knowledge incorporated in pre-trained BERT embeddings.
%Second, S-GCN can also handle typos, incompleteness and some variations in company names: "AxpoSolution" was linked correctly to "Axpo Holding AG" despite the missing space between the words; "GebIritY" was matched correctly to "Geberit AG" even with two errors in the input name; "Tetra Pak" was linked correctly to "Tetra Laval Holding \& Finance SA", showing that S-GCN distinguished that "Tetra" is the most important term in this task.
A video that demonstrates the examples discussed in this section and many other capabilities of the model is available at \url{https://youtu.be/LDzCZNrCm3o}.

%%%%%%%%%%%%%%%%%%%%%%%%%%%%%%%%%%%%%%%%%%%%%%%%%%%%%%%%%%%%%%%%%%%%%%%%%%%%%%%%%

%\section{Conclusion}
%\label{sec:conclusion}
%
%We have investigated how graph modeling and graph neural networks can be leveraged for data integration. We proposed a general GNN-based framework for learning hidden representations of entities.
%Experimental results show that GNNs and specifically S-GCNs are effective in performing business entity matching. %The key takeaways are: \textit{(i)} it is important to model the graph structure and node features and \textit{(ii)} using graph neural networks can improve data integration with structured and unstructured sources.
%Future work will aim at exploring several aspects that are both methodological and architectural.  On the methods side, we aim at providing guidance for cases where attributes in a record are better represented as node features rather than separate nodes.
%%, and when instead better performance can be obtained by creating a graph representation from them. 
%In addition, we will explore different training strategies and different GNN architectures.

\balance
\bibliographystyle{named}
\bibliography{references}

\end{document}